\pdfoutput=1

\documentclass[11pt]{article}

\usepackage{acl}

\usepackage{times}
\usepackage{latexsym}

\usepackage[T1]{fontenc}

\usepackage[utf8]{inputenc}

\usepackage{microtype}
\usepackage{epsfig,color,graphicx}
\usepackage{booktabs}

\usepackage{amsmath, multicol}

\title{Efficient slot labelling}

\author{Vladimir Vlasov \\
  at the time of working on this paper \\
  V.Vlasov worked at Cognigy GmbH, \\
  Germany}

\begin{document}
\maketitle
\begin{abstract}

Slot labelling is an essential component of any dialogue system, aiming to find important arguments in every user turn. Common approaches involve large pre-trained language models (PLMs) like BERT or RoBERTa, but they face challenges such as high computational requirements and dependence on pre-training data.
In this work, we propose a lightweight method which performs on par or better than the state-of-the-art PLM-based methods, while having almost 10x less trainable parameters. This makes it especially applicable for real-life industry scenarios.

\end{abstract}

\section{Introduction}

Task-oriented dialogue systems can assist users in attaining a well-defined goal: booking a restaurant \cite{multiwoz} or querying a banking issue \cite{banking77}. Slot labelling is an essential component of any such system, aiming to find important arguments in every user turn. For instance, in a restaurant booking dialogue system, the slot values \textit{Pizza Hut} and \textit{noon} from an input will be matched to \textit{name of restaurant} and \textit{time of booking} slots, respectively.

Constructing large datasets for every domain is a complex and costly process. Many efforts have been undertaken to offer methods to serve more resource-lean scenarios \cite{intentsslotfewshot, spanconvertfewshot}, when only a few labeled examples in a given domain are available. The solutions frequently rely on large pre-trained language models (PLMs) such as BERT \cite{bert} or RoBERTa \cite{xlmroberta}. There are several issues with these methods, chiefly: \textbf{a)} The models have high computational requirements which are not always available when the model is deployed in industrial settings, especially for on-device use; \textbf{b)} the success of PLMs is heavily reliant on the data used in pre-training \cite{convert}. In other words, when the user enters a value (e.g., \textit{Hackethal's}) that has not been seen in pre-training, the model might have difficulty in matching it to the right slot (e.g., \textit{restaurant}).

In this work, we propose a lightweight method solving the two aforementioned issues. Its performance is on par or better than state-of-the-art PLM-based methods, while having almost 10x less trainable parameters.

\section{Related work}

Parameter efficiency is an important topic for production applications. Usually efficiency is achieved by using a pre-trained large language model and introducing adapters that are tuned on downstream tasks \cite{transferfornlp1, adapterhub, ansell2022composable}.
In this work we show that specifically designed architecture performs better or on par with tuned models without the requirement to use large pre-trained models. Our work in this respect is more similar to classical word embeddings models where cross-domain knowledge was embedded into a small number of parameters~\citep{word2vec, fasttext, surveycrosslingual}.

\section{Methodology}

\subsection{Model}

Consider a task where we need to extract two types of entities \textit{from city} and \textit{to city}, e.g. \textit{I want to fly from [Moscow](from city) to [Berlin](to city)}. In order to differentiate between these two slot types, the model should pay attention to the functional words preceding the city names, namely, \textit{from} and \textit{to} in our example.
However, standard self-attention layer will calculate the score between the actual city names like "Moscow" or "Berlin" and functional words like "to" or "from". For the scores to carry the relevant information, it is desirable to include all combinations of the functional words and proper names, which is impossible, especially in production, as we cannot predict all values the users will approach the model with.

We propose to use a trainable embedding representing any word as a query in an attention mechanism. This way, the network learns to have high scores for words following specific functional words. Self-attention does not allow for this as it is impossible to differentiate between preceding and following words within the embeddings. To this end, we resort to relative attention.

The attention mechanism is prone to "memorize" that certain words correspond to certain entities, rather than trying to infer the entity from the context only. To better this, we apply masking to prevent the model to pay attention to tokens at current positions. However, depending on the entity type, the model sometimes needs to know actual word at current positions, so we add dense layer with sigmoid activation serving as a gate mechanism between attention output and word-level embedding.

\section{Experimental setup}

\subsection{Architecture}

The full architecture is composed of several layers.

\begin{figure*}[t]
\centering
\includegraphics[width=0.8\linewidth]{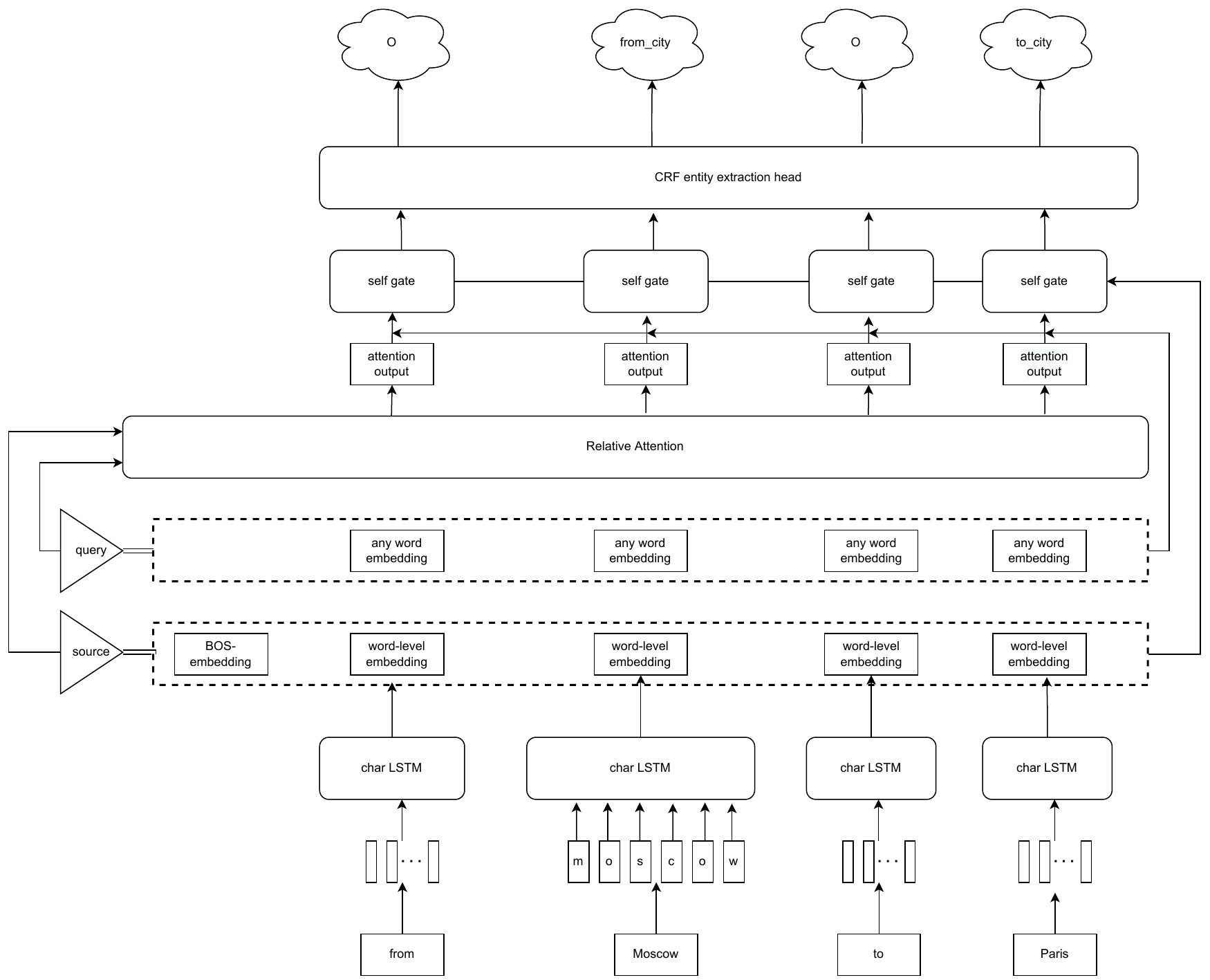}
\caption{The scheme of the model}
\label{fig:architecture}
\end{figure*}

The first layer is a character level long short-term memory (LSTM). The last output of the LSTM is passed through a dense layer, and the output of the dense layer serves as word-level embeddings. The next layer is the attention mechanism, followed by the gate that regulates whether the output of attention or actual word should be used for prediction. The last layer is CRF tagging that predicts the entities. The full scheme of the architecture is shown in Figure~\ref{fig:architecture}.

Mosig~and~Vlasov~\cite{sparselayers} observed that 80\% of the weights in a dense layer can be set to $0$ without any loss of accuracy for the downstream task. In this work, the relative sparsity was achieved by generating a random mask and applying it to the kernel of a dense layer. Such a method has one deficiency in that it doesn't reduce effective number of weights stored in the memory. In prior work \citet{blockdiagonals1, blockdiagonals2}, the authors show that a trained neural network can be compressed by rearranging weights into a block diagonal matrix by permutating the input units.

Here we introduce a block diagonal dense layer that is trained with block diagonal matrix of weights, and only the trainable weights are stored in the memory.
The eq.~\ref{eq:W} shows a block diagonal weight matrix with 2 blocks.
\begin{equation}
    W = \left[\begin{array}{ccccc}
        w_{11} & \cdots & w_{1n} & 0 & \cdots \\
        \vdots & \ddots & \vdots & \vdots & \ddots \\
        w_{m1} & \cdots & w_{mn} & 0 & \cdots \\
        0 & \cdots & 0 & w_{(m+1)(n+1)} & \cdots \\
        \vdots & \ddots & \vdots & \vdots & \ddots \\
        0 & \cdots & 0 & w_{(2m)(n+1)} & \cdots \\
    \end{array}\right]
    \label{eq:W}
\end{equation}

We can split input into two blocks as well.
\begin{equation}
    X = \left[\begin{array}{cccccc}
        x_{11} & \cdots & x_{1m} &  x_{1(m+1)} & \cdots & x_{1(2m)} \\
        x_{21} & \cdots & x_{2m} &  x_{2(m+1)} & \cdots & x_{2(2m)} \\
        \cdots & \cdots & \cdots & \cdots & \cdots & \cdots \\
    \end{array}\right]
    \label{eq:X}
\end{equation}
Matrix multiplication can be expressed as concatenated matrix multiplications of blocks.
\begin{equation}
\begin{gathered}
    XW = \left[\begin{array}{cc}
        X_1 & X_2
    \end{array}\right]
    \left[\begin{array}{cc}
        W_1 & 0 \\
        0 & W_2 \\
    \end{array}\right] \\
     = \left[\begin{array}{cc}
        X_1W_1 & X_2W_2
    \end{array}\right]
    \label{eq:XW}
\end{gathered}
\end{equation}

The eq.~\ref{eq:XW} can be implemented by the Einstein summation layer with equation $bkm,kmn->bnk$ with two additional reshape operations to split input into blocks before applying Einstein summation and merge blocked output after the Einstein summation layer.

\subsection{Hyperparameters}
We use the following hyper parameters:
number of units in attention mechanism = 256,
number of units in character lstm = 128,
attention head size = 128,
number of attention heads = 4,
number of dense diagonal blocks = 8,
dropout rate = 0.1,
attention dropout rate = 0.1,
weight decay rate = 0.01.

\subsection{Datasets}

We consider three datasets: RESTAURANTS-8K, MTOP and ATIS-en

\paragraph{RESTAURANTS-8K \cite{spanconvertfewshot}} The dataset
 contains conversations from a commercial restaurant booking system. It contains five entity types: date, time, people, first name, last name. The authors measure performance for smaller training subsets maintaining the same test set. Full training set size is 8128 examples.

 \paragraph{MTOP \cite{mtop}} The dataset consists of six languages: English, Spanish, French, German, Hindi and Thai. It contains 78 entity types across 11 domains. The total number of examples is 100k.

 \paragraph{ATIS \citep{hemphill-etal-1990-atis}} The dataset consists of annotated transcripts of audio recordings of people making flight reservations. The training, development, and test sets contain 4478, 500 and 893 utterances. It contains 79 entities.

\subsection{Baselines}

For our baselines:

Datasets RESTAURANTS-8K, MTOP, and ATIS-en use results from~\citet{spanconvertfewshot} in table~\ref{tab:restaurant8k}, ~\citet{mtop} in table~\ref{tab:mtop}, and \citet{xu2020end} in table~\ref{tab:atis}, respectively. Where applicable, baseline performance is taken from mentioned papers\footnote{F1 score in this work is calculated excluding non-entity "O" labels}.

\section{Results and Discussion}

We compare our results with baselines described above.
The results for RESTAURANTS-8K dataset are presented in table~\ref{tab:restaurant8k}. Results show our model achieves on par or stronger results compared with Span-ConveR, a significantly larger pretrained model. It is worth noting that our model brings especially large improvements in low-resource scenarios.

\begin{table}[h]
\centering
\resizebox{0.35\textwidth}{!}{%
{\footnotesize
\begin{tabular}{@{}ccc@{}}
\toprule
\textbf{Fraction} & \textbf{Span-ConveRT} & \textbf{Our Model} \\ \midrule
1   (8198)        & 0.96                  & 0.957              \\
1/2 (4099)        & 0.94                  & 0.949              \\
1/4 (2049)        & 0.91                  & 0.934              \\
1/8 (1024)        & 0.89                  & 0.912              \\
1/16 (512)        & 0.81                  & 0.866              \\
1/32 (256)        & 0.64                  & 0.753              \\
1/64 (128)        & 0.58                  & 0.599              \\
1/128 (64)        & 0.41                  & 0.484              \\
1/256 (32)        & ----                  & 0.361              \\ \bottomrule
\end{tabular}%
}
}
\caption{Average F1 scores over all slots for RESTAURANTS-8K dataset depending on training set fractions. Numbers in brackets represent training set sizes. First column represents performance of Span-ConveRT~\citet{spanconvertfewshot}. Second column is our model.}
\label{tab:restaurant8k}
\end{table}

Table~\ref{tab:mtop} shows results on MTOP dataset compared to pretrained XLM-R model. Except for English language our model performs better, while on English dataset it performs on par.

\begin{table}[h]
\centering
\resizebox{0.35\textwidth}{!}{%
{\footnotesize
\begin{tabular}{@{}ccc@{}}
\toprule
\textbf{Language} & \textbf{XLM-R} & \textbf{our F1 no O} \\ \hline
 en & 0.928 & 0.917 \\
 es & 0.899 & 0.913 \\
 fr & 0.883 & 0.893 \\
 de & 0.880 & 0.895 \\
 hi & 0.875 & 0.881 \\
 th & 0.854 & 0.891 \\ \bottomrule
\end{tabular}%
}
}
\caption{Average F1 scores over all slots for MTOP dataset for different languages. First column represents performance of XLM-R~\citet{mtop}. Second column is our model.}
\label{tab:mtop}
\end{table}

On ATIS dataset our model performs on par with pre trained BERT model~\ref{tab:atis}.

\begin{table}[h]
\centering
\resizebox{0.35\textwidth}{!}{%
{\footnotesize
\begin{tabular}{@{}ccc@{}}
\toprule
\textbf{Language} & \textbf{BERT} & \textbf{our F1 no O} \\ \hline
 en & 0.955 & 0.950 \\ \bottomrule
\end{tabular}%
}
}
\caption{Average F1 scores over all slots for ATIS dataset. First column represents performance of BERT~\citet{xu2020end}. Second column is our model.}
\label{tab:atis}
\end{table}



\subsection{Ablation Studies}


To analyze which components of the architecture contribute the most we run the ablation study. Results are presented in tables~\ref{tab:ablation-restaurant8k},~\ref{tab:ablation-mtop},~\ref{tab:ablation-atis}.

\begin{table*}[h]
\centering
\resizebox{0.65\textwidth}{!}{%
{\footnotesize
\begin{tabular}{@{}cccc@{}}
\toprule
 \textbf{Fraction} & \textbf{CRF F1 no O} & \textbf{Self Rel Attn F1 No O} & \textbf{Self Attn F1 No O} \\  \hline
 1   (8198) & 0.918 & 0.957 & 0.956 \\
 1/2 (4099) & 0.909 & 0.951 & 0.946 \\
 1/4 (2049) & 0.893 & 0.936 & 0.916 \\
 1/8 (1024) & 0.861 & 0.911 & 0.863 \\
 1/16 (512) & 0.776 & 0.864 & 0.801 \\
 1/32 (256) & 0.641 & 0.732 & 0.630 \\
 1/64 (128) & 0.583 & 0.634 & 0.534 \\
 1/128 (64) & 0.479 & 0.526 & 0.450 \\
 1/256 (32) & 0.481 & 0.469 & 0.318 \\ \bottomrule
\end{tabular}%
}
}
\caption{Average F1 scores over all slots for RESTAURANTS-8K dataset depending on training set fractions. Numbers in brackets represent training set sizes. First column represents performance of only CRF. Second column is the model with self relative attention. Third column is the model with self attention.}
\label{tab:ablation-restaurant8k}
\end{table*}

\begin{table}[h]
\centering
\resizebox{0.4\textwidth}{!}{%
{\footnotesize
\begin{tabular}{@{}ccc@{}}
\toprule
 \textbf{Language} & \textbf{CRF F1 no O} &  \textbf{Self Rel Attn F1 no O} \\ \hline
 en & 0.705 & 0.917 \\
 es & 0.706 & 0.914 \\
 fr & 0.668 & 0.894 \\
 de & 0.709 & 0.892 \\
 hi & 0.662 & 0.883 \\
 th & 0.661 & 0.892 \\ \bottomrule
\end{tabular}%
}
}
\caption{Average F1 scores over all slots for MTOP dataset for different languages. First column represents performance of only CRF. Second column is the model with self relative attention.}
\label{tab:ablation-mtop}
\end{table}

\begin{table*}[!ht]
\centering
\resizebox{0.63\textwidth}{!}{%
{\footnotesize
\begin{tabular}{@{}ccccc@{}}
\toprule
 \textbf{Language} & \textbf{Our F1 No O} & \textbf{CRF F1 No O} & \textbf{Self Rel Attn F1 no O} & \textbf{Self Attn F1 No O} \\ \\  \hline
 en        & 0.950 & 0.744 & 0.952 & 0.948 \\
 city tags & 0.869 & 0.521 & 0.844 & 0.843 \\ \bottomrule
\end{tabular}%
}
}
\caption{Average F1 scores over all slots for ATIS dataset. First column represents performance of our model. Second column is CRF only model. Third column is the model with self relative attention. Forth column is the model with self attention.}
\label{tab:ablation-atis}
\end{table*}

First of all we remove attention model and analyse the performance of pure CRF model. Based on the F1 scores we conclude that attention helps for different entity types. On RESTAURANTS-8K the drop is $~0.04$ while on MTOP and ATIS it is $~0.2$. We explain that difference in performance comes from the fact that MTOP and ATIS entities require more knowledge of the context than entities in RESTAURANTS-8K dataset.

We also substitute our attention module with self attention to analyze how much abstract query contributes to the performance. For all the datasets we can see that there is no drop of performance. However, in real production systems entity values that needs to be extracted are not present in the training set. To evaluate such performance we substitute cities in ATIS dataset with cities that are not present during training. We picked city because it is the largest category in ATIS and it has more than 1 context dependent type. On such dataset the performance drop is $0.025$.

On RESTAURANTS-8K when self relative attention is substituted with normal self attention we can see that the performance drops when training data is small: starting from $1/8$ the difference is about $0.05$.

\subsection{Efficiency}

One of the main properties of our model is its parameter efficiency. We achieve at least a 100X reduction in comparison to standard PLMs: our full model contains about $1$ mln parameters, cf. mBERT, XLM-R and XLM-R-Large with $178$, $270$ and $550$ mln parametrrs, respectively \cite{chung2020rethinking}. In addition, block diagonal layer with $8$ blocks brings further 4x parameter reduction  (app. \ref{sec:appendix trainable params}).

\section{Conclusion}

In this paper we presented an efficient entity extraction model with low number of trainable parameters. We showed that our model performs better or on par with large pretrained language models on different languages. Our ablation study shows how various architectual decisions contribute to the overall performance of the model. In order to further decrease amount of trainable parameters, we developed new dense layer with sparse block diagonal matrix and showed that using this layer doesn't reduce the performance.

\section*{Acknowledgements}

I would like to thank the team in Cognigy for providing me the opportunity to work on this paper.
Special thank to Sasha Khizov for creating an example of a chatbot where movie titles could not be extracted by already existing algorithms. I would also like to thank Evgeniia Razumovskaia for the discussions about the paper. I would also like to express my gratitude to Tristan Rayner for editing and proofreading the paper.

\bibliography{paper}

\begin{thebibliography}{20}
\expandafter\ifx\csname natexlab\endcsname\relax\def\natexlab#1{#1}\fi

\bibitem[{Ansell et~al.(2022)Ansell, Ponti, Korhonen, and
  Vuli{\'c}}]{ansell2022composable}
Alan Ansell, Edoardo Ponti, Anna Korhonen, and Ivan Vuli{\'c}. 2022.
\newblock Composable sparse fine-tuning for cross-lingual transfer.
\newblock In \emph{Proceedings of the 60th Annual Meeting of the Association
  for Computational Linguistics (Volume 1: Long Papers)}, pages 1778--1796.

\bibitem[{Bojanowski et~al.(2017)Bojanowski, Grave, Joulin, and
  Mikolov}]{fasttext}
Piotr Bojanowski, Edouard Grave, Armand Joulin, and Tomas Mikolov. 2017.
\newblock Enriching word vectors with subword information.
\newblock \emph{Transactions of the association for computational linguistics},
  5:135--146.

\bibitem[{Budzianowski et~al.(2018)Budzianowski, Wen, Tseng, Casanueva, Ultes,
  Ramadan, and Gasic}]{multiwoz}
Pawe{\l} Budzianowski, Tsung-Hsien Wen, Bo-Hsiang Tseng, I{\~n}igo Casanueva,
  Stefan Ultes, Osman Ramadan, and Milica Gasic. 2018.
\newblock Multiwoz-a large-scale multi-domain wizard-of-oz dataset for
  task-oriented dialogue modelling.
\newblock In \emph{Proceedings of the 2018 Conference on Empirical Methods in
  Natural Language Processing}, pages 5016--5026.

\bibitem[{Casanueva et~al.(2020)Casanueva, Tem{\v{c}}inas, Gerz, Henderson, and
  Vuli{\'c}}]{banking77}
I{\~n}igo Casanueva, Tadas Tem{\v{c}}inas, Daniela Gerz, Matthew Henderson, and
  Ivan Vuli{\'c}. 2020.
\newblock Efficient intent detection with dual sentence encoders.
\newblock In \emph{Proceedings of the 2nd Workshop on Natural Language
  Processing for Conversational AI}, pages 38--45.

\bibitem[{Chung et~al.(2020)Chung, Fevry, Tsai, Johnson, and
  Ruder}]{chung2020rethinking}
Hyung~Won Chung, Thibault Fevry, Henry Tsai, Melvin Johnson, and Sebastian
  Ruder. 2020.
\newblock Rethinking embedding coupling in pre-trained language models.
\newblock \emph{arXiv preprint arXiv:2010.12821}.

\bibitem[{Conneau et~al.(2020)Conneau, Khandelwal, Goyal, Chaudhary, Wenzek,
  Guzm{\'a}n, Grave, Ott, Zettlemoyer, and Stoyanov}]{xlmroberta}
Alexis Conneau, Kartikay Khandelwal, Naman Goyal, Vishrav Chaudhary, Guillaume
  Wenzek, Francisco Guzm{\'a}n, {\'E}douard Grave, Myle Ott, Luke Zettlemoyer,
  and Veselin Stoyanov. 2020.
\newblock Unsupervised cross-lingual representation learning at scale.
\newblock In \emph{Proceedings of the 58th Annual Meeting of the Association
  for Computational Linguistics}, pages 8440--8451.

\bibitem[{Coope et~al.(2020)Coope, Farghly, Gerz, Vuli{\'c}, and
  Henderson}]{spanconvertfewshot}
Samuel Coope, Tyler Farghly, Daniela Gerz, Ivan Vuli{\'c}, and Matthew
  Henderson. 2020.
\newblock Span-convert: Few-shot span extraction for dialog with pretrained
  conversational representations.
\newblock In \emph{Proceedings of the 58th Annual Meeting of the Association
  for Computational Linguistics}, pages 107--121.

\bibitem[{Devlin et~al.(2019)Devlin, Chang, Lee, and Toutanova}]{bert}
Jacob Devlin, Ming-Wei Chang, Kenton Lee, and Kristina Toutanova. 2019.
\newblock Bert: Pre-training of deep bidirectional transformers for language
  understanding.
\newblock In \emph{Proceedings of the 2019 Conference of the North American
  Chapter of the Association for Computational Linguistics: Human Language
  Technologies, Volume 1 (Long and Short Papers)}, pages 4171--4186.

\bibitem[{Hemphill et~al.(1990)Hemphill, Godfrey, and
  Doddington}]{hemphill-etal-1990-atis}
Charles~T. Hemphill, John~J. Godfrey, and George~R. Doddington. 1990.
\newblock \href {https://aclanthology.org/H90-1021} {The {ATIS} spoken language
  systems pilot corpus}.
\newblock In \emph{Speech and Natural Language: Proceedings of a Workshop Held
  at Hidden Valley, {P}ennsylvania, June 24-27,1990}.

\bibitem[{Henderson et~al.(2020)Henderson, Casanueva, Mrk{\v{s}}i{\'c}, Su,
  Wen, and Vuli{\'c}}]{convert}
Matthew Henderson, I{\~n}igo Casanueva, Nikola Mrk{\v{s}}i{\'c}, Pei-Hao Su,
  Tsung-Hsien Wen, and Ivan Vuli{\'c}. 2020.
\newblock Convert: Efficient and accurate conversational representations from
  transformers.
\newblock In \emph{Findings of the Association for Computational Linguistics:
  EMNLP 2020}, pages 2161--2174.

\bibitem[{Houlsby et~al.(2019)Houlsby, Giurgiu, Jastrzebski, Morrone,
  De~Laroussilhe, Gesmundo, Attariyan, and Gelly}]{transferfornlp1}
Neil Houlsby, Andrei Giurgiu, Stanislaw Jastrzebski, Bruna Morrone, Quentin
  De~Laroussilhe, Andrea Gesmundo, Mona Attariyan, and Sylvain Gelly. 2019.
\newblock Parameter-efficient transfer learning for nlp.
\newblock In \emph{International Conference on Machine Learning}, pages
  2790--2799. PMLR.

\bibitem[{Hsu et~al.(2020)Hsu, Chang, Shen, Shuai, Tseng, and
  Yang}]{blockdiagonals2}
Yun-Jui Hsu, Yi-Ting Chang, Chih-Ya Shen, Hong-Han Shuai, Wei-Lun Tseng, and
  Chen-Hsu Yang. 2020.
\newblock On minimizing diagonal block-wise differences for neural network
  compression.
\newblock In \emph{ECAI 2020}, pages 1198--1206. IOS Press.

\bibitem[{Li et~al.(2021)Li, Arora, Chen, Gupta, Gupta, and Mehdad}]{mtop}
Haoran Li, Abhinav Arora, Shuohui Chen, Anchit Gupta, Sonal Gupta, and Yashar
  Mehdad. 2021.
\newblock Mtop: A comprehensive multilingual task-oriented semantic parsing
  benchmark.
\newblock In \emph{Proceedings of the 16th Conference of the European Chapter
  of the Association for Computational Linguistics: Main Volume}, pages
  2950--2962.

\bibitem[{Mikolov et~al.(2013)Mikolov, Sutskever, Chen, Corrado, and
  Dean}]{word2vec}
Tomas Mikolov, Ilya Sutskever, Kai Chen, Greg~S Corrado, and Jeff Dean. 2013.
\newblock Distributed representations of words and phrases and their
  compositionality.
\newblock \emph{Advances in neural information processing systems}, 26.

\bibitem[{Mosig and Vlasov(2021)}]{sparselayers}
Johannes Mosig and Vladimir Vlasov. 2021.
\newblock Why rasa uses sparse layers in transformers.
\newblock https://rasa.com/blog/why-rasa-uses-sparse-layers-in-transformers/.
\newblock Accessed: 2024-01-16.

\bibitem[{Nesky and Stout(2018)}]{blockdiagonals1}
Amy Nesky and Quentin~F Stout. 2018.
\newblock Neural networks with block diagonal inner product layers.
\newblock In \emph{International Conference on Artificial Neural Networks},
  pages 51--61. Springer.

\bibitem[{Pfeiffer et~al.(2020)Pfeiffer, R{\"u}ckl{\'e}, Poth, Kamath,
  Vuli{\'c}, Ruder, Cho, and Gurevych}]{adapterhub}
Jonas Pfeiffer, Andreas R{\"u}ckl{\'e}, Clifton Poth, Aishwarya Kamath, Ivan
  Vuli{\'c}, Sebastian Ruder, Kyunghyun Cho, and Iryna Gurevych. 2020.
\newblock Adapterhub: A framework for adapting transformers.
\newblock In \emph{Proceedings of the 2020 Conference on Empirical Methods in
  Natural Language Processing: System Demonstrations}, pages 46--54.

\bibitem[{Ruder et~al.(2019)Ruder, Vuli{\'c}, and
  S{\o}gaard}]{surveycrosslingual}
Sebastian Ruder, Ivan Vuli{\'c}, and Anders S{\o}gaard. 2019.
\newblock A survey of cross-lingual word embedding models.
\newblock \emph{Journal of Artificial Intelligence Research}, 65:569--631.

\bibitem[{Xu et~al.(2020)Xu, Haider, and Mansour}]{xu2020end}
Weijia Xu, Batool Haider, and Saab Mansour. 2020.
\newblock End-to-end slot alignment and recognition for cross-lingual nlu.
\newblock In \emph{Proceedings of the 2020 Conference on Empirical Methods in
  Natural Language Processing (EMNLP)}, pages 5052--5063.

\bibitem[{Yu et~al.(2021)Yu, He, Zhang, Du, Pasupat, and
  Li}]{intentsslotfewshot}
Dian Yu, Luheng He, Yuan Zhang, Xinya Du, Panupong Pasupat, and Qi~Li. 2021.
\newblock Few-shot intent classification and slot filling with retrieved
  examples.
\newblock In \emph{Proceedings of the 2021 Conference of the North American
  Chapter of the Association for Computational Linguistics: Human Language
  Technologies}, pages 734--749.

\end{thebibliography}
\bibliographystyle{acl_natbib}

\appendix
\label{sec:appendix}

\section{Number of Trainable Parameters for Different Datasets}
\label{sec:appendix trainable params}

\begin{table}[h]
\centering
\resizebox{0.3\textwidth}{!}{%
{\footnotesize
\begin{tabular}{@{}cccc@{}}
\toprule
 \textbf{Language} & \textbf{Full Dense} & \textbf{8 Block Dense} & \textbf{Reduc. Factor}\\ \hline
en & 1023132 & 253084 & 4.04 \\ \bottomrule
\end{tabular}%
}
}
\caption{Number of trainable parameters for ATIS dataset. First column represents our model but with full dense layers. Second column is our model with block diagonal dense layers.}
\label{tab:num-params-atis}
\end{table}

\begin{table}[!h]
\centering
\resizebox{0.3\textwidth}{!}{%
{\footnotesize
\begin{tabular}{@{}cccc@{}}
\toprule
 \textbf{Fraction} & \textbf{Full Dense} & \textbf{8 Block Dense} & \textbf{Reduc. Factor}\\ \hline
 1   (8198) & 999222 & 229174 & 4.37 \\
 1/2 (4099) & 998710 & 228662 & 4.36 \\
 1/4 (2049) & 998198 & 228150 & 4.38\\
 1/8 (1024) & 996662 & 226614 & 4.40 \\
 1/16 (512) & 995126 & 225078 & 4.42\\
 1/32 (256) & 994614 & 224566 & 4.43\\
 1/64 (128) & 993078 & 223030 & 4.45\\
 1/128 (64) & 990518 & 220470 & 4.49\\
 1/256 (32) & 987958 & 217910 & 4.53\\ \bottomrule
\end{tabular}%
}
}
\caption{Number of trainable parameters for RESTAURANTS-8K dataset depending on training set fractions. Numbers in brackets represent training set sizes. First column represents our model but with full dense layers. Second column is our model with block diagonal dense layers.}
\label{tab:num-params-restaurant8k}
\end{table}

\begin{table}[!h]
\centering
\resizebox{0.3\textwidth}{!}{%
{\footnotesize
\begin{tabular}{@{}ccc@{}}
\toprule
 \textbf{Language} & \textbf{Full Dense} & \textbf{8 Block Dense} \\ \hline
 en & 1027802 & 257754 \\
 es & 1031814 & 261766 \\
 fr & 1037122 & 267074 \\
 de & 1032922 & 262874 \\
 hi & 1059244 & 289196 \\
 th & 1059756 & 289708 \\ \bottomrule
\end{tabular}%
}
}
\caption{Number of trainable parameters on MTOP dataset for different languages. First column represents our model but with full dense layers. Second column is our model with block diagonal dense layers.}
\label{tab:num-params-mtop}
\end{table}

\section{Detailed Results for Different Setups}
\begin{table}[h]
\centering
\resizebox{0.3\textwidth}{!}{%
{\footnotesize
\begin{tabular}{@{}cc@{}}
\toprule
 \textbf{Fraction} &\textbf{Full Dense F1 No O} \\ \hline
 1   (8198) & 0.957 \\
 1/2 (4099) & 0.950 \\
 1/4 (2049) & 0.932 \\
 1/8 (1024) & 0.908 \\
 1/16 (512) & 0.860 \\
 1/32 (256) & 0.758 \\
 1/64 (128) & 0.637 \\
 1/128 (64) & 0.527 \\
 1/256 (32) & 0.364 \\ \bottomrule
\end{tabular}%
}
}
\caption{Average F1 scores over all slots for RESTAURANTS-8K dataset depending on training set fractions. Numbers in brackets represent training set sizes. First columns represents performance of our model but with full dense layers.}
\end{table}

\begin{table}[h]
\centering
\resizebox{0.3\textwidth}{!}{%
{\footnotesize
\begin{tabular}{@{}cc@{}}
\toprule
 \textbf{Language} &\textbf{Full Dense F1 No O} \\ \hline
  en & 0.9508\\ \bottomrule
\end{tabular}%
}
}
\caption{Average F1 scores over all slots for ATIS dataset. First columns represents performance of our model but with full dense layers.}
\end{table}

\begin{table}[h]
\centering
\resizebox{0.3\textwidth}{!}{%
{\footnotesize
\begin{tabular}{@{}cc@{}}
\toprule
 \textbf{Language} &\textbf{Full Dense F1 No O} \\ \hline
   en & 0.916 \\
 es & 0.911 \\
 fr & 0.888 \\
 de & 0.893 \\
 hi & 0.885 \\
 th & 0.890 \\ \bottomrule
\end{tabular}%
}
}
\caption{Average F1 scores over all slots for MTOP dataset for different languages. First column represents performance of our model but with full dense layers.}
\end{table}


\end{document}